\documentclass[letterpaper, 10 pt]{ieeetran} 
                                                         
\usepackage{amsmath,amssymb,amsfonts,epsf,epsfig,times}
\usepackage[all]{xy}
\usepackage{graphicx,color}
\usepackage{subfigure}
\usepackage{url}
\usepackage{cite}
\usepackage{tikz}
\usepackage{epstopdf}
\usepackage{algorithm}
\usepackage[noend]{algpseudocode}
\usepackage{leftidx}

\usepackage{tikz}
\usepackage{amsmath}

\DeclareMathOperator*{\argmin}{arg\,min}
\usepackage{makecell}

 \newcommand{\re}{\mathbb{R}}
 \newcommand{\Acal}{\mathcal{A}}

\title{\LARGE \bf
Graph Based Multi-layer K-means++ (G-MLKM) for Sensory Pattern Analysis in Constrained Spaces
}

\author{Feng Tao, Rengan Suresh, Johnathan Votion, and Yongcan Cao
\thanks{The authors are with the Department of Electrical and Computer Engineering, University of Texas, San Antonio, TX 78249, USA.} 
\thanks{Corresponding Author: Yongcan Cao (yongcan.cao@utsa.edu)}
}

\begin{document}

\maketitle
\thispagestyle{empty}
\pagestyle{empty}

\begin{abstract}
In this paper, we focus on developing a novel unsupervised machine learning algorithm, named graph based multi-layer k-means++ (G-MLKM), to solve data-target association problem when targets move on a constrained space and minimal information of the targets can be obtained by sensors. Instead of employing the traditional data-target association methods that are based on statistical probabilities, the G-MLKM solves the problem via data clustering. We first will develop the Multi-layer K-means++ (MLKM) method for data-target association at local space given a simplified constrained space situation. Then a p-dual graph is proposed to represent the general constrained space when local spaces are interconnected. Based on the dual graph and graph theory, we then generalize MLKM to G-MLKM by first understanding local data-target association and then extracting cross-local data-target association mathematically analyze the data association at intersections of that space. To exclude potential data-target association errors that disobey physical rules, we also develop error correction mechanisms to further improve the accuracy. Numerous simulation examples are conducted to demonstrate the performance of G-MLKM. 
\end{abstract}

\begin {keywords}
Graph theory, MLKM, clustering, data preprocessing, data-object association.
\end{keywords}

\section{Introduction}
Associating data with the right target in a multi-target environment is an important task in many research areas, such as object tracking \cite{vo2015multitarget}, surveillance \cite{benfold2011stable,haritaoglu2000w}, and situational awareness \cite{endsley1995toward}. Image sensors can be used to acquire rich information related to each target, which will significantly simplify the data-target association problem. For example, video cameras in a multi-target tracking mission can provide colors and shapes of targets as extra features in the association process~\cite{pasula1999tracking}. However, considering the costs, security issues, and special environments (e.g., ocean tracking~\cite{fortmann1983sonar}, military spying), a simple, reliable, and low-cost sensor network is often a preferred option~\cite{singh2007tracking}. Consequently, the data-target association problem needs to be further studied, especially in cases when the gathered data are cluttered and contains limited information related to the targets. 

The existing approaches for data-target association are, in general, consisted of three procedures~\cite{grisetti}: \textit{(i) Measurements collection} - preparation before data association process, such as object identification in video frames, radar signals processing, or raw sensor data accumulation; \textit{(ii) Measurements prediction} - predict the potential future measurements based on history data, which yields an area (validation gate) that narrows down the search space; \textit{(iii) Optimal measurement selection} - select the optimal measurement that matches history data according to a criterion (varies in different approaches) and update the history dataset. With the same procedures but different choices of the optimal measurement criteria, many data-target association techniques have already been developed. Among them, the well-known techniques include the global nearest neighbor standard filter (Global NNSF) \cite{konstantinova2003study}, joint probabilistic data association filter (JPDAF)  \cite{bar2011tracking, ma2006distributive,kim2016jpdas,yuhuan2014modified}, and multiple hypothesis tracking (MHT) \cite{Blackman04}.  

The Global NNSF approach attempts to find the maximum likelihood estimate related to the possible measurements (non-Bayesian) at each scan (that measures the states of all targets simultaneously). For nearest neighbor correspondences, there is always a finite chance that association is incorrect \cite{cox1993review}. Besides that, the Global NNSF assumes a fixed number of targets and cannot adjust the target number during the data association process. A different well-known technique for data association is JPDAF, which computes association probabilities (weights) and updates the track with the weighted average of all validated measurements. Similar to Global NNSF, JPDAF cannot be applied in scenarios with targets birth and death \cite{vo2015multitarget}. The most successful algorithm based on this data-oriented view is the MHT \cite{reid1979algorithm}, which takes a delayed decision strategy by maintaining and propagating a subset of hypotheses in the hope that future data will disambiguate decisions at present \cite{vo2015multitarget}. MHT is capable of associating noisy observations and is resistant to a dynamic number of targets during the association process. The main disadvantage of MHT is its computational complexity as the number of hypotheses increases exponentially over time.   

There are other approaches available for data association. For example, the Markov chain Monte Carlo data association (MCMCDA) \cite{pasula1999tracking, oh2004markov}. MCMCDA takes the data-oriented, combinatorial optimization approach to the data association problem but avoids the enumeration of tracks by applying a sampling method called Markov chain Monte Carlo (MCMC) \cite{oh2004markov}, which still implements statistical probabilities in the procedure of optimal measurement selection.  

The main contribution of this paper is the development of an efficient unsupervised machine learning algorithm, called Graph Based Multi-layer K-means++ (G-MLKM). The proposed G-MLKM differs from the existing data-target association methods in three aspects. First, in contrast to the previous developed data association approaches that estimate the potential measurement from history data for each target and select an optimal one from validated measurements based on statistical probabilities, G-MLKM solves the data-target association problem in the view of data clustering. Second, the previous approaches are mainly developed with respect to sensors that are capable of obtaining information from a multiple dimensional environment, such as radars, sonars, and video cameras. G-MLKM is proposed on sensors that only provide limited information. An interesting research on tracking targets with binary proximity sensors can be seen in \cite{singh2007tracking}, whose objective is only limited to target counting, while G-MLKM can associate data to targets. Third, G-MLKM can address the case that targets move in a constrained space, which requires dealing with data separation and merging. 

The reminder of this paper is structured as follows. The data association problem in a constrained space and the corresponding tasks are described in Section \ref{sec:PF}. In Section \ref{sec:MLKM}, the multi-layer k-means++ (MLKM) method is developed for data-target association at local space given a simplified constrained space situation. The graph based multi-layer k-means++ (G-MLKM) algorithm is then developed in Section \ref{sec:G-MLKM} for general constrained spaces. Simulation examples are then provided in Section \ref{sec:simu}. Section~\ref{sec:con} provides a brief summary of the work presented in this paper.

\section{Problem Formulation} \label{sec:PF}



In this paper, we consider the problem of data-target association when multiple targets move across a road network. Here, a road network is a set of connected road segments, along which low-cost sensors are spatially distributed. The sensors are used to collect information of targets, which, in particular, are the velocity of targets and the corresponding measured time. We assume 1) \textcolor{black}{there is no false alarm in the sensor measurements,} and 2) the target's velocity does not change rapidly within two adjacent sensors. The collected information about a target is normally disassociated with the target itself, meaning that the target from which the information was captured cannot be directly identified using the information. Hence, data-target associations is necessary.     

Fig. 1 shows \textcolor{black}{one road network example} that consists of 6 road segments. Without loss of generality, let the total number of road segments in one road network be denoted as $L$. The road segments are denoted as $R_1, R_2, \cdots, R_{L}$, respectively. The length of road segment $R_i$ is denoted as $D_i$ for $i = 1, 2, \cdots, L$. To simplify discussion, we assume the road segments are for one-way traffic, i.e., targets cannot change their moving directions within one road segment. However, when the road segment allows bidirectional traffic, we can separate it into two unidirectional road segments and the proposed approach in this paper directly applies. Let $\mathcal{S}_i=\{S_{i1}, S_{i2}, \cdots, S_{iN_{i}}\}$ be a set of $N_{i}\in \mathbb{R}$ sensors placed along the direction of road segment $R_i$. In other words, for sensor $S_{ij} \in \mathcal{S}_i$, the larger the sub-notation $j$ is, the further distance the sensor locates away from the starting point of road segment $R_i$. We denote the corresponding distance between sensor $S_{ij}$ and the starting point of road segment $R_i$ as $d_{ij}$. Hence, the position set for sensors in $R_i$ related to the starting point can be denoted as ${\mathcal{P}}_i =\{d_{i1}, d_{i2}, \cdots, d_{iN_i}\}$, where $0 \leq d_{i1} < d_{i2} < \cdots < d_{iN_i} \leq D_i$. 

\begin{figure}[t]
	\centering
	\includegraphics[width=0.6\linewidth]{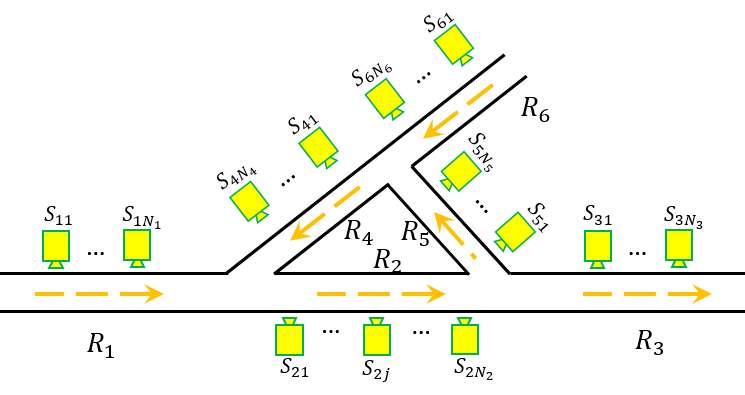}
	\caption{An example road network. $R_i$ represents the road segment. $S_{ij}$ represents the $j$th sensor on the $i$th road segment.}
	\label{fig:road_network}
\end{figure}


For each sensor $S_{ij}$, its measurements are collected and stored in a chronological order. The collections are denoted as a column vector $X_{ij}$, such that
$
X_{ij} = [\textbf{x}^1_{ij}, \textbf{x}^2_{ij}, \cdots, \textbf{x}^{m_{ij}}_{ij}]',
$
where $i \in \{1,$ $2, \cdots, L\}$, $j \in \{1,$ $2, \cdots, N_i\}$, the prime symbol represents the transpose operation for a vector, $m_{ij}$ is the total number of measurements in $X_{ij}$, and $\textbf{x}^n_{ij}$, $n \in \{1, 2, \cdots, m_{ij}\}$, denotes an individual measurement in $X_{ij}$. In particular, $\textbf{x}^n_{ij} = [v^n_{ij}, t^n_{ij}]$ stores the measured velocity $v^n_{ij}$ when one target passed by sensor $S_{ij}$ at time $t^n_{ij}$. As the elements in $X_{ij}$ are stored in the chronological order, the recorded time for each measurement satisfies $t^1_{ij} < t^2_{ij} < \cdots < t^{m_{ij}}_{ij}$, which can be distinguished based on the superscript $n$. All the measurement vectors stored by sensors that locate in the same road segment $R_i$ are stored into a matrix $\mathcal{X}_i$, such that
$
\mathcal{X}_i = [\bar{X}_{i1}, \bar{X}_{i2}, \cdots, \bar{X}_{i{N_i}}],
$
where $\mathcal{X}_i \in \mathbb{R}^{m_{i}\times N_i}$, $m_i = \textup{max}_j\{m_{ij}\}$, and the column of the matrix is defined as $\bar{X}_{ij} = [X_{ij}, \textbf{0}^{1\times (m_{i}-m_{ij})}]'$. If $m_{ij} = m_i$, $\bar{X}_{ij} = X_{ij}$. The added all-zero row vector in $\bar{X}_{ij}$ is to unify the length of vectors in matrix $\mathcal{X}_i$ considering that miss detection may happen or targets may remain (or stop) inside the road network for a given data collection period. 

The road network collects $\mathcal{X}_i, i=1,\cdots,L$ that only include information of target's velocity and the corresponding measurement time. In order to solve data-target association based on the $L$ matrices, three tasks need to be accomplished. The first task (\textit{Task 1}) is to cluster $\mathcal{X}_i$ into $m_i$ groups for each road segment. Denote the data grouping result for each road segment as a new matrix $\mathcal{T}_i$, such that 
\begin{equation}\label{eq: result for 1D}
\mathcal{T}_i = [\bar{T}_{i1}, \bar{T}_{i2}, \cdots, \bar{T}_{i{m_i}}]',\quad i = 1, 2, \cdots, L
\end{equation}
where $\bar{T}_{iz}, z = 1, 2, \cdots, m_i,$ is a row vector consisting of $N_i$ measurements associated with the same target, defined as
\begin{align}\label{eq4}
\bar{T}_{iz} = [\boldsymbol{\tau}^1_{iz}, \boldsymbol{\tau}^{2}_{iz}, \cdots, \boldsymbol{\tau}^{N_i}_{iz}],
\end{align}
where $\boldsymbol{\tau}^u_{iz}$ is an entry of $\bar{X}_{iu}$ for $u = 1, 2, \cdots, N_i$. Then a new row vector $T_{iz}$ is obtained from $\bar{T}_{iz}$ by excluding all zero elements.

The second task (\textit{Task 2}) is to link the trajectories of targets at road intersections by pairing sensor $S_{i1}/ S_{iN_i}$ from multiple road segments that are connected geometrically. In particular, let $O^{i_{nts}}_T$ denote the index set of road segments that have outgoing targets related to one intersection $i_{nts}$, and $I^{i_{nts}}_T$ denote the index set of road segments that have ingoing targets related to the same intersection. Since the road segments are unidirectional, the two index sets have no overlaps, i.e., $O^{i_{nts}}_T \cap I^{i_{nts}}_T = \emptyset$. Datasets that belong to targets that move towards the intersection ${i_{nts}}$ is denoted as 
\begin{equation}\label{eq5}
Q^{i_{nts}}_I = \{ \textbf{x}^{k_O}_{iN_i}| \; \forall \textbf{x}^{k_O}_{iN_i} \in X_{iN_i}, \forall i \in O^{i_{nts}}_T \},
\end{equation}
while datasets that belong to targets that leave the intersection ${i_{nts}}$ is denoted as 
\begin{equation}\label{eq6}
Q^{i_{nts}}_O = \{ \textbf{x}^{k_I}_{i1}| \; \forall \textbf{x}^{k_I}_{i1} \in X_{i1}, \forall i \in I^{i_{nts}}_T \}.
\end{equation}
where $k_O\in\{1, 2, \cdots, m_{iN_i}\}$, $k_I\in\{1, 2, \cdots, m_{i1}\}$, $O^{i_{nts}}_T \subset \{1, 2, \cdots, L\}$, and $I^{i_{nts}}_T \subset \{1, 2, \cdots, L\}$. Because targets may stop in the intersection or the data collection process terminates before targets exit the intersection, the total number of targets heading into an intersection ${i_{nts}}$ is always greater than or equal to the number of targets leaving the same intersection, i.e., $|Q^{i_{nts}}_I|\geq |Q^{i_{nts}}_O|$. For simplicity of notation, denote $|Q^{i_{nts}}_I|$ and $|Q^{i_{nts}}_O|$ as $n_I$ and $n_O$. Then we can calculate $n_I$ and $n_O$ via
\begin{equation}\label{eq: cardinality}
n_I  = \sum_{\forall i \in O^{i_{nts}}_T} m_{iN_i}\textup{ and }
n_O  = \sum_{\forall i \in I^{i_{nts}}_T} m_{i1}.
\end{equation}
The pairing task for intersection ${i_{nts}}$ can be denoted as a mapping function $f$, such that
\begin{equation}\label{eq7}
f(\textbf{x}^k_{iN_i}) \mapsto \textbf{x}_{l1}, \quad\forall k \in \{1, 2, \cdots, n_I\},
\end{equation}
where $\textbf{x}^k_{iN_i} \in Q^{i_{nts}}_I$ and $\textbf{x}_{l1} \in \{Q^{i_{nts}}_O, {0, 0, \cdots, 0}_{n_I - n_O}\}$. In particular, the function $f$ for intersection ${i_{nts}}$ can be denoted as a permutation matrix $G_{i_{nts}} \in \mathbb{R}^{n_I \times n_I}$. 

The last task (\textit{Task 3}) is to merge data groups on the road network when loops may exist, i.e., targets may pass the same road segment several times. Hence, multiple data association groups may belong to the same target. The merged results can be denoted as $L$ symmetric matrices $G_{R_i} \in$ $ \mathbb{R}^{m_i\times m_i}$ for each road segment $R_i$. If targets only pass the road segment $R_i$ once, $G_{R_i}$ is an identity matrix.         

In this paper, we are going to propose a new unsupervised machine learning algorithm to associate data-target for the collected $L$ matrices. In particular, this algorithm first creates a new clustering structure for data grouping in each matrix (associated with each road segment), and then leverages graph theory and clustering algorithms to link the matrices from different road segments for each intersection. Finally, the entire dataset can be analyzed and associated properly to the targets. The output of this new algorithm will be a detail trajectory path for each target with the captured velocities along the road segments. In the next two sections, the new data-target associations algorithm will be explained in detail. We begin the discussion with a special case when the road network is consisted of a single road segment.

\section{MLKM for a single road segment} \label{sec:MLKM}
\begin{figure}[!ht]
	\centering
	\includegraphics[width=0.5\linewidth]{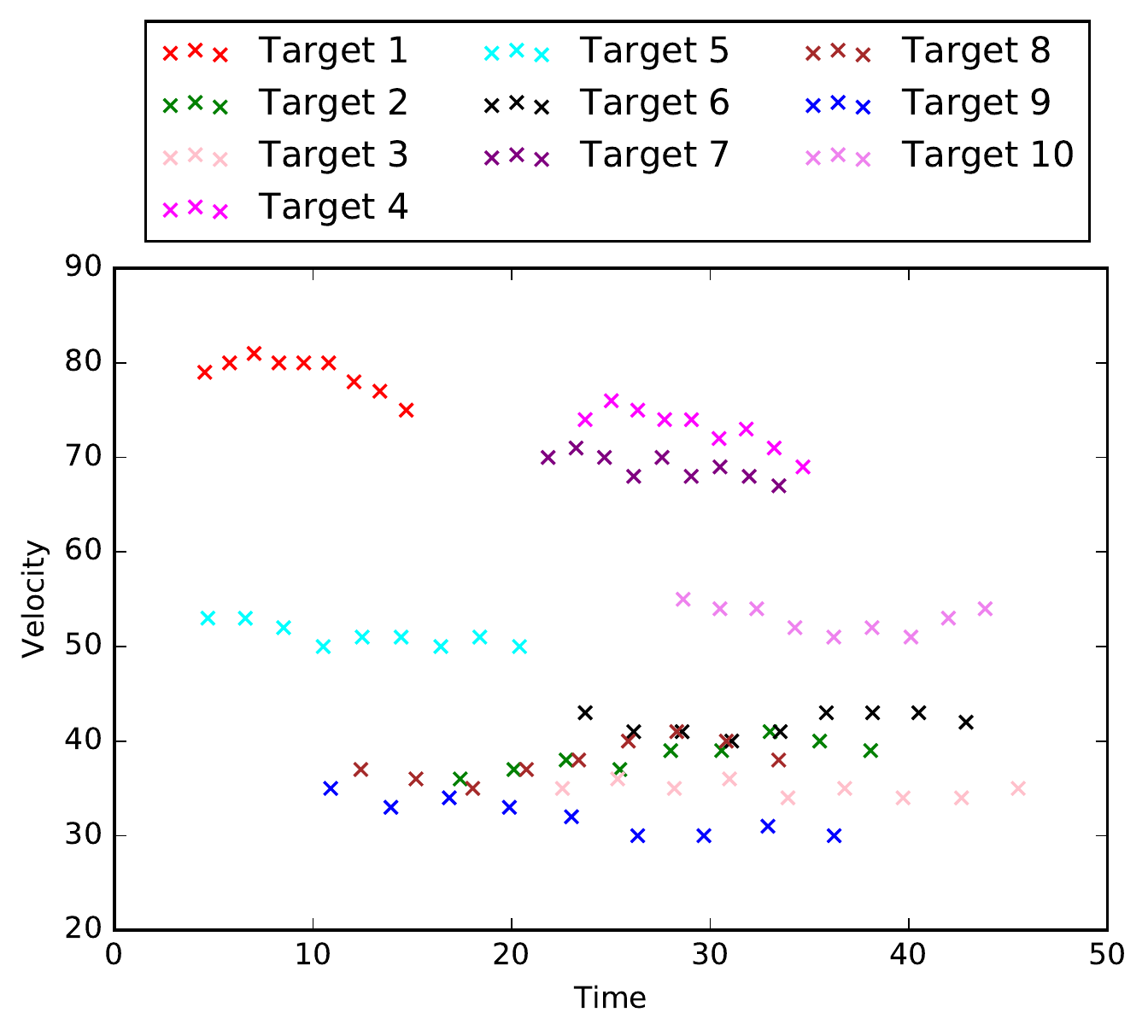}
	\caption{Example of $\mathcal{X}_1$ for a single road segment.}
	\label{fig:1D_dataset_example}
\end{figure}

In this section, we consider the special case when $L = 1$, i.e., the road network is only consisted of one road segment, $R_1$. In this special case, there are neither intersections nor loops in the road network. Therefore, the tasks in identifying data-target associations are simplified to cluster $\mathcal{X}_1$ into $m_1$ groups (\textit{Task 1}) only. One example of matrix $\mathcal{X}_1 \in \mathbb{R}^{10\times 9}$ is shown in Fig. \ref{fig:1D_dataset_example}, which is the plot of measurements for 10 different targets that are captured by 9 equally spaced sensors on road segment $R_1$. 

\subsection{K-means++ Clustering and Deep Neural Network} \label{section: k-means}
K-means~\cite{kanungo2002efficient} and k-means++ \cite{arthur2007k} are perhaps the most common methods for data clustering. For a set of data points in a $N$-dimensional system, the two algorithms perform clustering by grouping the points that are closer to the optimally placed centroids. From the machine learning perspective, k-means learns where to optimally place a pre-defined number of centroids such that the cost function, defined as
$
\Phi_Y(\mathcal{C}) = \sum_{y\in Y} d^2(y,\mathcal{C})\textup{,} \nonumber 
$
is minimized, where $d(y,\mathcal{C}) = {\min}_{y\in Y}\left \| y-c_i \right \|$ represents the distance between a sub-set of measurements $Y$ and a centroid $c_i$ and $\mathcal{C} = \{c_1,...,c_k\}$ represents the set of centroids.
The associated cost function is the sum of the Euclidean distances from all data points to their closer centroid. The cost function and optimization algorithms are the same for k-means and k-means++ while the only difference between them is that k-means++ places the initial guesses for the centroids in places that have data concentration, and consequently improves the running time of Lloyd's algorithm and the quality of the final solution~\cite{arthur2007k}. 

\textcolor{black}{A much more complex boundary may exist between two data groups. Therefore, we also verify the potential performance of Deep Neural Network (DNN) algorithm~\cite{hinton2002neural} in the data association process, which is known for the capability of recognizing underlying patterns and defining better decision boundaries among data samples. For the purpose of evaluating the supervised DNN capabilities, a slight modification of the problem is considered. Instead of a complete unlabeled dataset $\mathcal{X}_1$, part of the measurements are pre-labeled, i.e., data-target relations for part of the measurements are known. Also, we extend the measurement's dimensions to further include $vt$, $v^2t$, and $vt^2$ as extra features so that the inner structure of DNN can be simpler. Table \ref{tb:parameters_nn} presents the detail settings of the DNN framework. }

\begin{table}[ht]
	\centering
	\caption{DNN configuration parameters.}
	\begin{tabular}{|c | c|}
		
		\hline 
		Framework & Definition \\ 
		\hline 
		Cost Function & Softmax  \\
		Activation Function & Relu  \\ 	
		Optimizer & Adam Optimizer  \\	 
		Number of Hidden Layers&  2\\      
        Number of Neurons &  8\\ 
		\hline 
	\end{tabular}
	\label{tb:parameters_nn}%
\end{table}

\subsection{K-means++ with data preprocessing}\label{subsect: preprocessing}
While DNN can potentially provide better performance for the data association problem, it demands labeled datasets for training. In real scenarios, however, the training dataset may not be available. In contrast, k-mean++ can cluster data samples without the need for labeled dataset. This unsupervised property of k-means++ enables a wider application domain. Hence, k-means++ is more practical for the task of clustering $\mathcal{X}_1$ into $m_1$ groups. Moreover, when the dataset $\mathcal{X}_1$ is small and sparse, k-means++ can perform well on the task of data-target association.      

However, when the measurements are distributed along the time axis and velocity profiles are close, k-means++ tends to place the centroids in positions where data from different targets that overlap and hence causes an inaccurate data-target pairing. This happens because k-means implements Euclidean distance to determine which centroid data sample $(v,t)$ belongs, i.e., 
\begin{equation}\label{eq:euclidean}
\argmin_{(v^*_i, t^*_i) \in \mathcal{C}} \sqrt{(v-v^*_i)^2+(t-t^*_i)^2},
\end{equation}
where $\mathcal{C}$ is the set of centroids. When data samples distribute along time axis, the time difference becomes the determining factor for grouping results.

\begin{figure}[!ht]
	\centering
	\includegraphics[width=0.5\linewidth]{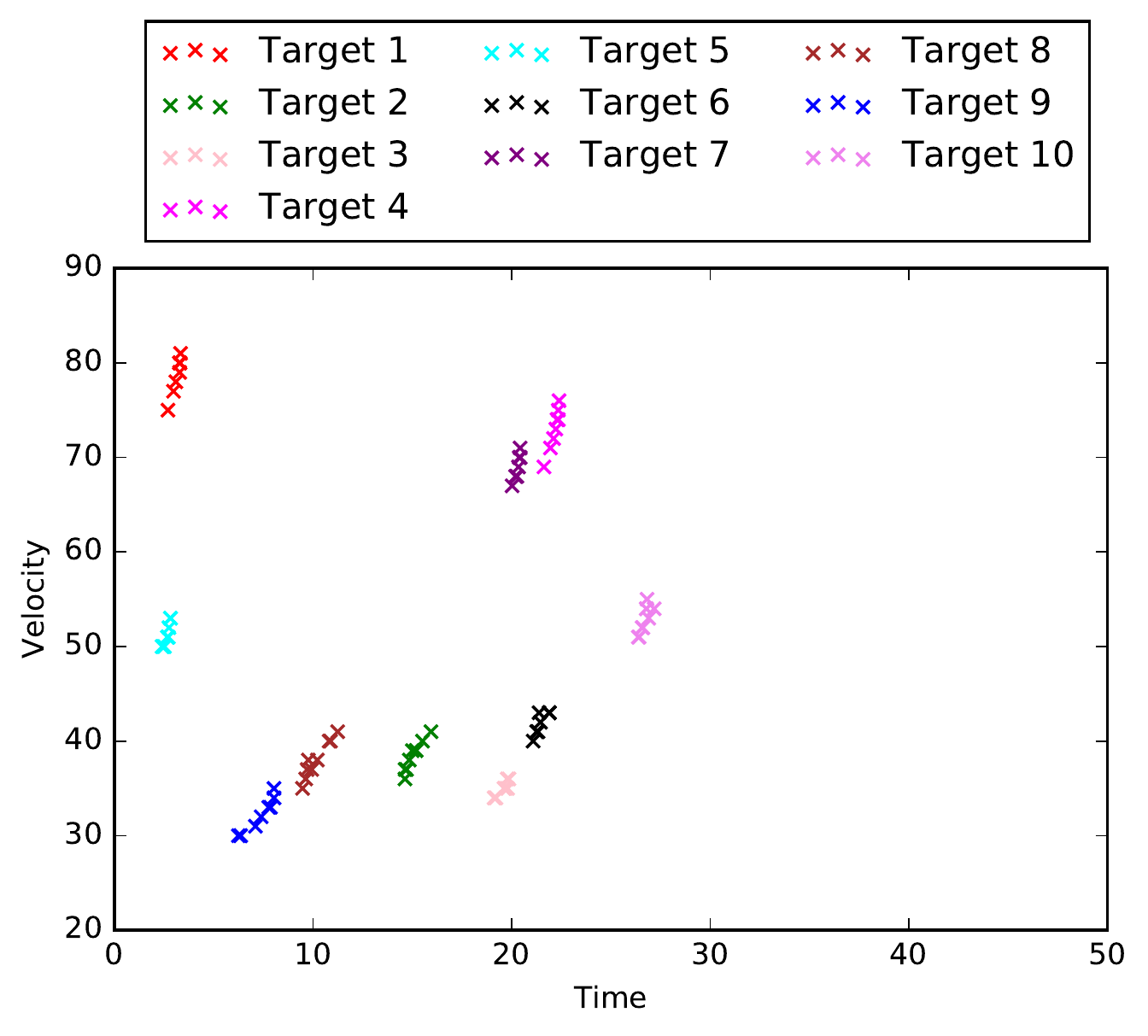}
	\caption{Example of preprocessed $\mathcal{X}_1$ for a single road segment.}
	\label{fig:1D_dataset_pre_example}
\end{figure}

One natural way to balance the two components (time difference and velocity difference) in~(\ref{eq:euclidean}) is to process $\mathcal{X}_1$ before applying k-means++. The idea of preprocessing is similar to the principal component analysis~\cite{einasto2011sdss} that projects data into a main axis. The preprocessed data sample is denoted as $\hat{\textbf{x}}^n_{1j} = [v^n_{1j},\; \hat{t}^n_{1j}]$, where $\hat{t}^n_{1j}$ is given by 
\begin{equation}\label{eq: preprocessing}
\hat{t}^n_{1j} = t^n_{1j} - \frac{d_{1j}- d^*}{v^n_{1j}},  
\end{equation}
where $j \in \{1, \cdots, N_1\},~n \in \{ 1, \cdots, m_{1j}\}$, $d_{1j}$ is the position of sensor $S_{1j}$ with respect to the starting point of road segment $R_1$, and $d^*$ is the reference point for projecting. Fig.~\ref{fig:1D_dataset_pre_example} is the preprocessed result for the dataset in Fig. \ref{fig:1D_dataset_example}. In this example, the reference point $d^*$ is select to be the starting point, and we can see clusters for each target have been formed after data preprocessing.

\subsection{Multi-layer K-means++}
Through the preprocessing procedure, data can be roughly separated for different targets that provide dense and grouped subsets. The boundaries between two groups, however, maybe still too complex for k-means++ to define, especially, when $\mathcal{X}_1$ is a large dataset and the grouped subsets are close to each other. Inspired by the DNN capability of defining classification boundaries via a multi-layer structure and a back-propagation philosophy, we propose a new multi-layer k-means++ (MLKM) method that integrates the DNN's multi-layer structure with the clustering capabilities of k-means++ to overcome the complex boundary challenge. 


The proposed MLKM algorithm is performed via 3 layers: \textit{(i) - data segmentation and clustering} - The dataset is sequentially partitioned into smaller groups for the purpose of creating sparse data samples for k-means++; \textit{(ii) - Error detection and correction} - Check the clustered data by searching for errors through predefined rules and re-cluster the data using nearest neighbor concepts~\cite{arya1994optimal} if an error is found. Note that the k-means++ associates the data closer to the optimally placed centroid based on the Euclidean distance between data point and centroid, which is a scalar quantity; \textit{(iii) - Cluster matching} - match the clusters of each segment by preprocessing the cluster centroids of all segments to the cluster centroid of the first segment and again grouping them based on k-means++. A detail explanation for these three layers are given as follows.


\subsubsection*{Layer 1 \textit{(Data Segmentation \& Clustering)}}

Without loss of generality, we assume that there are $K$ sensors per segment. The dataset $\mathcal{X}_1\in \mathbb{R}^{m_1 \times N_1}$ ($m_1$ and $N_1$ are the maximum number of measurements and the total sensor number in sensor set $\mathcal{S}_1$, respectively) is sequentially partitioned into $E$ segments, such that 
\begin{align*}
E = \left\{
\begin{array} {ll}
N_1/K,&N_1\%K=0,\\
N_1/K+1,&\text{otherwise}.
\end{array}\right.
\end{align*} 
In other words, when $N_1\%K\neq 0$, the last segment will contain measurements from less than $K$ sensors. In the following of the paper, we assume that $N_1\%K=0$ in the following of this paper for the simplicity of presentation. When $N_1\%K\neq0$, we can add some extra artificial sensors with all zero measurements. Then the data segment can be defined as 
$
\mathcal{X}_{1e}= \bigcup_{j=(e-1)K+1}^{eK} \bar{X}_{1j},
$
where $e =1, 2, \cdots, E$. K-means++ algorithm is then applied to each $\mathcal{X}_{1e}$ by excluding all zero elements. By aggregating the clustering results, we can obtain a set of centroids for $\mathcal{X}_{1e}$, $e =1, 2, \cdots, E$, defined as ${\mathcal{C}}_{1e} = \{c_{11}^e,c_{12}^e,\cdots,c_{1m_1}^e\},$ and the associated measurements with each $c_{1k}^e$ centroid are represented as \textcolor{black}{${T}_{1k}^e$}, where $k \in \{1, 2, \cdots, m_1\}$.

\subsubsection*{Layer 2 \textit{(Error Detection \& Correction)}} \label{subsection: error correction}
The first layer seeks to associate data for each data segment. Because the clustering standard used in k-means++ is a scalar quantity while the actual measurements are given by vectors, there are potential data association errors in \textcolor{black}{${T}_{1k}^e$}. Hence an additional layer to perform error detection and correction is needed. The error detection is to verify logic rules to determine if wrong data association appears in \textcolor{black}{${T}_{1k}^e$}. The error correction will conduct data re-association on the identified wrong associations. To avoid the same wrong reassociation again, the global nearest neighbor standard is chosen as the re-association technique instead of k-means++ given the assumption that the target's velocity does not change rapidly within two adjacent sensors. 

We here proposed the following logic rules for error detection:
\begin{itemize}
	\item $|\textcolor{black}{{T}_{1k}^e| > K}$;
	\item $\exists n_1 \neq n_2$, $\textbf{x}^{n_1}_{1l} \in \textcolor{black}{{T}_{1k}^e}$, $\textbf{x}^{n_2}_{1j} \in \textcolor{black}{{T}_{1k}^e}$  $\Rightarrow l=j$;
	\item $\exists l \geq j$, ${\textbf{x}^{n_1}_{1l}}_{\ne 0} \in \textcolor{black}{{T}_{1k}^e}$, $\textbf{x}^{n_2}_{1j} \in \textcolor{black}{{T}_{1k}^e}$  $\Rightarrow t^{n_1}_{1l} \leq t^{n_2}_{1j}$;
\end{itemize}
where $|\textcolor{black}{{T}_{1k}^e}|$ indicates the cardinality of $\textcolor{black}{{T}_{1k}^e}$. The first rule means that \textcolor{black}{more than $K$ measurements appear in ${T}_{1k}^e$.} The second rule means that more than one sensory measurements from the same sensor are associated with one target in $\bar{T}_{1k}^e$. The third rule means that target is recorded in a later time by a previous sensor. If one or more rules are satisfied, the corresponding $\textcolor{black}{{T}_{1k}^e}$ is then considered to be an erroneous data association and will be stored in $\mathcal{Y}^{*}_{1e}$, where $\mathcal{Y}^{*}_{1e}$ refers to the wrong data associations in $\mathcal{X}_{1e}$. 

The error correction is to re-associate data in $\mathcal{Y}^{*}_{1e}$ for the purpose of breaking all the logic rules listed above. We propose to use the global nearest neighbor approach. Specifically, elements in $\mathcal{Y}^{*}_{1e}$ that belongs to measurements of sensor $S_{1\ell}$ are selected sequentially to be evaluated against with every measurement in $\mathcal{Y}^{*}_{1e}$ that belongs to measurements of sensor $S_{1(\ell+1)}$ to obtain the best match. The evaluation is accomplished via the following optimization process:   
\begin{align}
\arg \min_{\kappa} ~& t^\kappa_{1(\ell+1)}-\left(t_{1\ell} + \frac{\left \| d_{1(\ell+1)}-d_{1\ell} \right \|}{v_{1\ell}}\right), \nonumber\\
~\textup{s.t.} ~ &\textbf{x}^\kappa_{1(\ell+1)} \in \mathcal{Y}^{*}_{1e}. \nonumber 
\end{align}
With this procedure, all $\textcolor{black}{{T}_{1k}^e}$ are updated with the corrected clusters and all $c_{1k}^e$ are re-calculated based on the updated $\textcolor{black}{{T}_{1k}^e}$. The new corrected set of centroids $\mathcal{C}_{1e}$ is updated for all segments and grouped into $\mathcal{C}_1 =\{\mathcal{C}_{11}~\mathcal{C}_{12}~...~\mathcal{C}_{1E} \}$. The position of the centroid set $\mathcal{C}_{1e}$ is defined as
\begin{equation}\label{eq: centroid_position}
\mathsf{d}_{1e} = \sum\limits_{j = (e-1)K +1}^{eK} d_{1j}/K.
\end{equation}

\subsubsection*{Layer 3 \textit{(Cluster Matching)}}
Through the preceding two layers, data-target association can be accomplished for each data segment $\mathcal{X}_{1e}$ independently. However, the target associations are uncorrelated among each data segment. In particular, the unsupervised k-means++ only groups data samples that belong to the same target while the clusters of each target are anonymous. Hence, it is still unclear how to associate the clusters among different segments. 

In Layer 3, we project $\mathcal{C}_{1e},~e=1,\cdots,E,$ using the preprocessing technique that is stated in~\ref{subsect: preprocessing}. More precisely, the time component in $c_{1k}^e \in \mathcal{C}_{1e}$ is preprocessed as 
\begin{equation*}\label{eq: MLKM-preprocessing}
\hat{t}_{1k}^e = t_{1k}^e - \frac{\mathsf{d}_{1e}- \mathsf{d}_{11}}{v_{1k}^e},\forall e \in \{1, \cdots, E\}, \forall k \in \{ 1, \cdots, m_{1}\}
\end{equation*}
where $c_{1k}^e = [v_{1k}^e,  t_{1k}^e]$, and $\mathsf{d}_{1e}$ is the position of centroid set $\mathcal{C}_{1e}$ defined in (\ref{eq: centroid_position}). Then k-means++ is applied to the preprocessed $\mathcal{C}_1$ to find the clusters that group cluster centroids in different data segments. Accordingly, the associated measurements $\textcolor{black}{{T}_{1k}^e}$ with respect to each centroid are merged together as $\textcolor{black}{{T}_{1k}}$ and, hence, provides the complete data-target association result for the entire road segment.

Note that the proposed MLKM method may not be applied directly to the case when $L > 1$ (i.e., more than one road segments). Therefore, we propose a more general method, named G-MLKM, to solve the general data-target association problem for a general road network in the next section. 

\section{G-MLKM for a general road network} \label{sec:G-MLKM}
In this section, we consider the general case when the road network is consisted of multiple road segments. To solve the data-target association problem, we propose a new graph-based multi-layer k-means++ (G-MLKM) algorithm. In particular, G-MLKM uses graph theory to represent the road network as a graph, and then links data from different road segments at each intersection of the road network by analyzing the graph structure. The data-target association problem for a general road network is then solved by merging the clustering results at intersections with the MLKM results on each road segment.  

We first briefly introduce graph theory and the representation of road networks using graphs as preliminaries. Then the procedures for G-MLKM are explained in detail. In particular, we begin with a new graph representation for the road network. Then the procedures for linking measurements at intersections (\textit{Task 2}) are described. After that, we unify the results on road segments and intersections, and complete the data merging task (\textit{Task 3}). 

\subsection{Preliminaries}
\subsubsection{Graph Theory}
For a system of $L$ connected agents, its network topology can be modeled as a directed graph $\mathcal{G} = (\mathcal{V}, \mathcal{E})$, where $\mathcal{V} = \{\mathtt{v}_1, \mathtt{v}_2, \cdots, \mathtt{v}_L\}$ and $\mathcal{E} \subseteq \mathcal{V} \times \mathcal{V}$ are, respectively, the set of agents and the set of edges that connect the agents. An edge $(\mathtt{v}_i, \mathtt{v}_j)$ in set $\mathcal{E}$ means that the agent $\mathtt{v}_j$ can access the state information of agent $\mathtt{v}_i$, but not necessarily \textit{vice versa}~\cite{cao2013overview}. The adjacency matrix $\mathcal{A} \in \mathbb{R}^{L\times L}$ of the directed graph $\mathcal{G}$ is defined by $\Acal=[a_{ij}]
\in\re^{L\times L}$,
where $a_{ij} =  1$ if $(\mathtt{v}_i, \mathtt{v}_j) \in \mathcal{E}$ and $a_{ij}=0$ otherwise.

\subsubsection{Graph Representation of Road Networks}
There are mainly two strategies to represent road networks using graph, namely primal graph and dual graph~\cite{zhao2015statistical}. In a primal graph representation, road intersections or end points are represented by agents and road segments are represented by edges~\cite{porta2006network}, while in a dual graph representation, road segments are represented by agents and an edge exists if two roads are intersected with each other~\cite{porta2006network2}. Compared with primal graph, dual graph concerns more on the topological relationship among road segments. As the data-target associations for each road segment can be solved by the MLKM method, the focus here is to cluster data at each intersection. As a consequence, the dual graph is a better option. However, the geometric properties such as road length are neglected by dual graph. Hence, some further modification to the dual graph is needed.

\begin{figure}[t]
	\centering
	\includegraphics[width=0.8\linewidth]{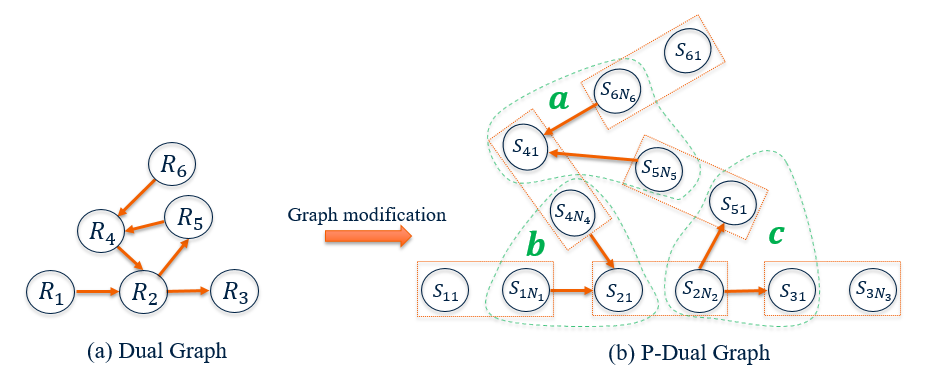}
	\caption{(a) Dual graph representation for the road network in Fig.~\ref{fig:road_network} with the nodes and arrows representing, respectively, the agents and the directed edges. (b) P-dual graph representation for the road network in Fig.~\ref{fig:road_network}, where two sensor nodes represent one road segment and the edge within the two sensor nodes are ignored. In this example, there exist 3 subgraphs which are denoted as $a, b,$ and $c$.}
	\label{fig:graph}
\end{figure}


\subsection{G-MLKM Algorithm}

In this subsection, we will provide the detail procedures for the G-MLKM algorithm that are composed of the following three steps.

\subsubsection{Modified Graph Representation for Road Networks}
Considering the cases when targets may stop in a road segment or data collection process may terminate before targets  pass through a road segment, the total number of measurements collected by sensor $S_{iN_i}$ (locates near the ending point of road segment $R_i$) may be less than the one collected by sensor $S_{i1}$ (locates near the starting point of road segment $R_i$). If the entire road segment is abstracted as one single agent, the inequality of measurements in the road segment may create issues for the subsequent data-target associations process. Here, we modify the dual graph by incorporating the primal graph for the representation of the road segment. In other words, we propose to replace each road segment node in the dual graph by two agents with one directed edge connecting them and the direction of the edge is determined by the traffic direction. In particular, we use the sensor nodes $S_{i1}$ and $S_{iN_i}$ as the two agents. We may neglect the edge between $S_{i1}$ and $S_{iN_i}$ because we focus on data-target associations at intersections while the data-target associations within the road segment can be accomplished by the MLKM method without the need for the knowledge of the graph. Moreover, the connection between $S_{i1}$ and $S_{iN_i}$ is unidirectional when the traffic is unidirectional. We call the new graph ``p-dual graph'', i.e., prime-based dual graph. An example of how to derive the p-dual graph is shown in Fig.~\ref{fig:graph}, where the original 6 agents in the dual graph are replaced by 12 agents and the edges between $S_{i1}$ and $S_{iN_i}$ are removed in the p-dual graph.

For a general road network with $L$ edge segments, the edges of the new p-dual graph is given by $\mathcal{V}^*=$ $\{S_{11},$ $S_{1N_1},$ $S_{21},$ $\cdots,$ $S_{L1},$ $S_{LN_L}\}$ with the corresponding adjacency matrix, $\mathcal{A}^*\in \mathbb{R}^{2L\times 2L}$, given by  
\begin{align}\label{eq: adjac}
\Acal^*=[a^*_{ij}]\in\re^{L\times L}, \quad a^*_{ij} =  
\begin{bmatrix}
0 & 0\\
a_{ij} & 0\\
\end{bmatrix}.
\end{align}


\subsubsection{Graph Analysis for Data Pairing at Intersections}
From $\mathcal{A}^*$ defined in~(\ref{eq: adjac}), we can observe that the adjacency matrix $\mathcal{A}^*$ has $L$ columns and $L$ rows that are all zeros.
Hence, the sparse matrix $\mathcal{A}^*$ can be further analyzed and decomposed to extract subgraphs related to different intersections. Then the task of linking the trajectories of targets at road intersections can be equivalently solved via pairing measurements of sensor $S_{i1}/S_{iN_i}$ from road segments in the subgraphs, which is further decomposed into the following three procedures.   

\begin{algorithm}[t!]
	\caption{Subgraph Extraction}\label{alg:graph-reduce}
	\begin{algorithmic}[1]
		\State{Input: $ \forall b_{ij} \in \mathcal{A}^*$;}
		\State{Output: $(O^{i_{nts}}_T, I^{i_{nts}}_T)$, ${i_{nts}} \in \{a, b, c, \cdots\}$}
        \State{$\textup{Idx}_{row}$ = $\textup{Idx}_{col}$ = $\{1, 2, \cdots, |\mathcal{A}^*|\};$}
        \State{$i$ = 0;} 
        \For{$i_{nts}$ in $\{a, b, c, \cdots \}$}
        \State{$O^{i_{nts}}_T = I^{i_{nts}}_T = \emptyset;$}
        \If{$|\textup{Idx}_{row}| \geq 1$}
        \Procedure{Increment}{$i$}
        \State{$i$ = $i$ + $1$;}
        \If{$i \in $ $\textup{Idx}_{row}$}
        \State{\textbf{return }$i$;}
        \Else
        \State{\textsc{Increment}{($i$)};}
        \EndIf
        \EndProcedure
        
        \Procedure{Recursion}{$i$}        
        \If{$\sum\nolimits_{\forall j \in \textup{Idx}_{col}} b_{ij} \geq 1$}
        \State{$O^{i_{nts}}_T = O^{i_{nts}}_T \cup \left\{i\right\};$}
        \Procedure{Extract}{$i$}
        \For{$j$ in $\textup{Idx}_{col}$}
        \If{$b_{ij} \neq 0$}
        \State{$I^{i_{nts}}_T = I^{i_{nts}}_T \cup \left\{j\right\};$}
        \EndIf
        \EndFor
        \State{$\textup{Idx}_{col}$ = $\textup{Idx}_{col}$$\setminus I^{i_{nts}}_T;$}
        \State{$\textup{Idx}_{row}$ = $\textup{Idx}_{row}$$\setminus \{i\};$}       
        \For{$j \in I^{i_{nts}}_T$}
        \If{$\sum\nolimits_{\forall l \in \textup{Idx}_{row}} b_{lj} \geq 1$}
        \For{$l$ in $\textup{Idx}_{row}$}
        \If{$b_{lj} \neq  0$}
        \State{$O^{i_{nts}}_T = O^{i_{nts}}_T \cup \left\{l\right\};$}
        \EndIf
        \EndFor
        \EndIf
        \EndFor
        \If{$\textup{Idx}_{row} \cap O^{i_{nts}}_T \neq \emptyset$}
        \State $\textsc{Extract}(\exists l \in (\textup{Idx}_{row} \cap O^{i_{nts}}_T))$;
        \Else
        \State{\textbf{return} $(O^{i_{nts}}_T, I^{i_{nts}}_T)$};
        \EndIf
        \EndProcedure
        \Else
        \State{$\textup{Idx}_{row}$ = $\textup{Idx}_{row}$$\setminus \{i\};$}
        \State {$i$ = $i$ + $1$;}
        \State $\textsc{Recursion}(i)$;
        \EndIf
        \EndProcedure
        \Else
        \State {\textbf{break};}
		\EndIf
        \EndFor
	\end{algorithmic}
\end{algorithm}

\subsubsection*{i.\,Subgraph Extraction}
The first procedure is to extract subgraphs from $\mathcal{A}^*$. Let the letters in alphabet $\{a, b, c, ...\}$ denote the names for different intersections. The subgraph extraction procedure begins with an intersection name as $a$, follows by $b$, $c$, and so on. For any intersection $i_{nts}$, the subgraph extraction is conducted by cross-searching the non-zero entries of the matrix $\mathcal{A}^*$ in a repeated row and column pattern. The corresponding indices of row and column containing non-zero entries, indicating the agents and edges that are included in that subgraph, are stored in the sets  $O^{i_{nts}}_T$ and $I^{i_{nts}}_T$, respectively. More precisely, $O^{i_{nts}}_T$ denotes the index set of road segments that have outgoing targets related to intersection $i_{nts}$ and $I^{i_{nts}}_T$ denotes the index set of road segments that have ingoing targets related to the same intersection. The index storing processes are defined as
$
O^{i_{nts}}_T = O^{i_{nts}}_T \cup \left\{i\right\},
$
and
$
I^{i_{nts}}_T = I^{i_{nts}}_T \cup \left\{j\right\},
$
where $i,j$ are the corresponding row index and column index, respectively. The iterative search process will terminate and return $(O^{i_{nts}}_T, I^{i_{nts}}_T)$ when there is no more non-zero element in the recorded row and column indices. 
Algorithm~\ref{alg:graph-reduce} is the pseudo code for the subgraph extraction procedure. The extracted results are denoted as $(O^{i_{nts}}_T, I^{i_{nts}}_T)$, where ${i_{nts}} \in \{a, b, c, \cdots\}$.


\subsubsection*{ii.\,Data Preprocessing at Intersections}
Given the subgraph that describes an intersection $i_{nts}$ is available from the preceding subgraph extraction procedure, datasets of $X_{i1}/X_{iN_i}$ which are subjected to the pairing task for the corresponding intersection can be pinpointed. In particular,~(\ref{eq5}) and (\ref{eq6}) define the dataset for the intersection $i_{nts}$ as an incoming dataset $Q^{i_{nts}}_I$ and an outgoing dataset $Q^{i_{nts}}_O$, respectively. As we assume that 1) no false alarm in the measurements, and 2) the target's velocity does not change rapidly within two adjacent sensors, data pairing at intersections may interpret as data clustering. A potential machine learning technique for data clustering is the k-means++. However, the sensors $S_{i1}/S_{iN_i}$ from different road segments are not guaranteed to locate near each other for a road intersection, which may contribute to a relatively large time difference in two sensors' measurements for one target. Hence, before applying k-means++, data preprocessing on $Q^{i_{nts}}_I$ and $Q^{i_{nts}}_O$ is necessary. 

\begin{figure}[t]
	\centering
	\includegraphics[width=0.6\linewidth]{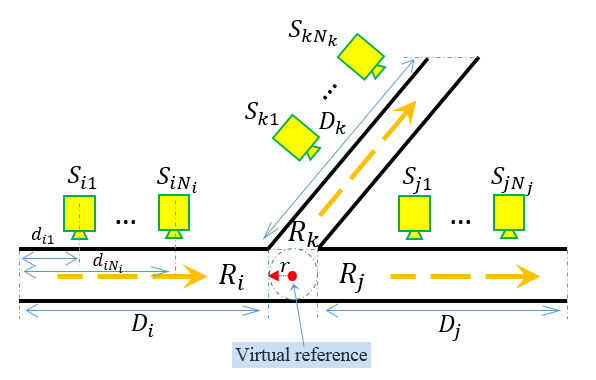}
	\caption{An intersection consists of three road segments denoted as $R_i$, $R_j$, and $R_k$. The virtual reference for data preprocessing is in the center of the intersection with a radius of $r$ to each road segment ending point.}
	\label{fig:intersection}
\end{figure}


Based on the proposed preprocessing definition in (\ref{eq: preprocessing}), we here propose a new data preprocessing technique that first selects a virtual reference at the center of the intersection $i_{nts}$ and then recomputes $\hat{t}^k_{ij}$ via projecting each element in $I^{i_{nts}}_T$ and $O^{i_{nts}}_T$ to the virtual reference as
\begin{equation}\label{eq: preprocessing2} 
\hat{t}^k_{ij} =
\begin{cases}
t^k_{i1} - \frac{r + d_{i1}}{v^k_{i1}}, & i\in I^{i_{nts}}_T \,\& \, j = 1\\ 
t^k_{iN_i} + \frac{D_i - d_{iN_i} + r}{v^k_{iN_i}}, & i\in O^{i_{nts}}_T \,\& \, j = N_i
\end{cases},
\end{equation}
where $k\in\{ 1, 2, \cdots, m_{ij}\}$ and $r$ is the radius of the intersection circle centered at the virtual reference. An example of locating the virtual reference is shown in Fig. \ref{fig:intersection}, where the intersection is consisted of 3 road segments denoted as $R_i, R_j$ and $R_k$. 

\subsubsection*{iii.\,Data Pairing at Intersections and Error Correction}
Denote the preprocessed datasets for $Q^{i_{nts}}_I$ and $Q^{i_{nts}}_O$ as $\hat{Q}^{i_{nts}}_I$ and $\hat{Q}^{i_{nts}}_O$. Then k-means++ can be applied to the preprocessed intersection datasets $\{\hat{Q}^{i_{nts}}_I, \hat{Q}^{i_{nts}}_O\}$ for data pairing. Similar to the development of MLKM for the case of one road segment, errors may arise when conducting the data pairing/clustering. Error detection and correction are needed to further improve the accuracy. 

For an intersection $i_{nts}$, the cardinalities of the preprocessed $\hat{Q}^{i_{nts}}_I$ and $\hat{Q}^{i_{nts}}_O$ remain the same as those of $Q^{i_{nts}}_I$ and $Q^{i_{nts}}_O$. As defined in (\ref{eq: cardinality}), 
$
|\hat{Q}^{i_{nts}}_I| = n_I \textup{ and } |\hat{Q}^{i_{nts}}_O| = n_O,
$
where $n_I \geq n_O$. 
The set of centroids is denoted as
$
{\mathcal{C}}_{i_{nts}} = \{c_{i_{nts}1},c_{i_{nts}2},\cdots,c_{i_{nts}n_I}\},
$
and the associated measurements with each centroid $c_{i_{nts}j}$, $j\in \{1, 2, \cdots, n_I\}$, are given as $Y_{i_{nts}j}$. The error correction is similar to the Layer 2 in the MLKM method described in section \ref{subsection: error correction}, and defines three logic rules for error detection: 
\begin{itemize}
	\item $|Y_{i_{nts}j}| > 2$;
	\item $|Y_{i_{nts}j} \cap \hat{Q}^{i_{nts}}_I| \ne 1$;
	\item $\textbf{x}_{iN_i} \in Y_{i_{nts}j}, {\textbf{x}_{l1}}_{\ne 0} \in Y_{i_{nts}j} \Rightarrow t_{l1} \leq t_{iN_i}$;
\end{itemize}
where $|Y_{i_{nts}j}|$ is the cardinality of $Y_{i_{nts}j}$. The first rule means more than two measurements are associated in $Y_{i_{nts}j}$. Error can be determined in this case because each target has at most two measurements in one intersection. The second rule means either none or more than one sensory measurements can be found from the incoming dataset $\hat{Q}^{i_{nts}}_I$. The third rule means that the outgoing measurement in $Y_{i_{nts}j}$ is recorded earlier than the incoming measurement. If one or more rules are satisfied, the corresponding $Y_{i_{nts}j}$ is then considered to be an erroneous data association and will be stored in $\mathcal{Y}_{i_{nts}}$. The error correction is to re-associate data in $\mathcal{Y}_{i_{nts}}$ for the purpose of breaking all the three logic rules listed above. To achieve this goal, we separate $\mathcal{Y}_{i_{nts}}$ into two subsets denoted as $\mathcal{Y}^I_{i_{nts}}$ and $\mathcal{Y}^O_{i_{nts}}$ given by
\begin{equation*}
\begin{aligned}
&\mathcal{Y}^I_{i_{nts}} = \{\textbf{x}_{iN_i}|\; \forall\textbf{x}_{iN_i} \in \mathcal{Y}_{i_{nts}}\}, \mathcal{Y}^O_{i_{nts}} = \{\textbf{x}_{l1}|\; \forall\textbf{x}_{l1} \in \mathcal{Y}_{i_{nts}}\},
\end{aligned}
\end{equation*}
where $\mathcal{Y}^I_{i_{nts}}$ and $\mathcal{Y}^O_{i_{nts}}$ store all measurements $\textbf{x}_{iN_i}$ and $\textbf{x}_{l1}$ in $\mathcal{Y}_{i_{nts}}$, respectively. Re-associate data in $\mathcal{Y}_{i_{nts}}$ becomes a linear assignment problem~\cite{kuhn1955hungarian} between $\mathcal{Y}^I_{i_{nts}}$ and $\mathcal{Y}^O_{i_{nts}}$. The optimal pairing between $\mathcal{Y}^I_{i_{nts}}$ and $\mathcal{Y}^O_{i_{nts}}$ can be found when the matching score reaches to the minimum via solving the optimization problem of
$
\arg \min_{M} ||M\times Y^I_{i_{nts}} - Y^O_{i_{nts}} ||, \nonumber 
$
where $Y^I_{i_{nts}}\in \mathbb{R}^{m_I\times 1}$ and $ Y^O_{i_{nts}}\in \mathbb{R}^{m_O\times 1}$ are column vectors converted from subsets $\mathcal{Y}^I_{i_{nts}}$ and $\mathcal{Y}^O_{i_{nts}}$, respectively. $M\in \mathbb{R}^{m_O\times m_I}$ is a special binary matrix with the summation of each row being 1. 

After the error correction is accomplished, all $Y_{i_{nts}j}$ will be updated to complete \textit{Task 2}. Furthermore, a permutation matrix $G_{i_{nts}} \in \mathbb{R}^{n_I \times n_I}$ can be created to record the pairing relationship between incoming dataset $Q^{i_{nts}}_I$ and outgoing dataset $Q^{i_{nts}}_O$ for each intersection. 

\subsubsection{Group Merging in the Road Network}
K-means++ clustering on the preprocessed dataset at each intersection solves the task of linking the trajectories of targets at road intersections \textit{(Task 2)} while the proposed MLKM method solves the task of data associations for each road segment \textit{(Task 1)}. If the clustering results at all intersections are combined with the MLKM results on all road segments, trajectory awareness for each target in the road network is achieved. This is valid for situations when targets only pass the same road segment once. However, when targets pass the same road segment and intersection for multiple times, one target can be assigned to multiple associated data groups on the road segment. To determine the connections among all associated data groups, an extra task (\textit{Task 3}) for merging data groups in the road network is needed. Given that the datasets at intersections are extracted from the $L$ matrices collected from all road segments, clusters at the intersections can be classified based on the data groups for all road segments. Therefore, the task of determining the connections among the associated data groups in the road network can be focused on connections of $\bar{T}_{iz}$ defined in~(\ref{eq4}) for each road segment. 

Let the symmetric matrix $G_{R_i} \in \mathbb{R}^{m_i\times m_i}$ denote the connections among the $m_i$ association groups in road segment $R_i$ given by $
G_{R_i} =[b_{ij}]\in\re^{m_i\times m_i}
$
where 
\begin{equation*}
b_{pq} = b_{qp} = 
\begin{cases}
1, \quad \textup{if } \bar{T}_{ip}, \bar{T}_{iq} \textup{ belong to the same target},\\
0, \quad \textup{otherwise}.
\end{cases}
\end{equation*}
To determine the entries in $G_{R_i}$, the depth-first search (DFS) \cite{tarjan1972depth} is implemented to detect cycles in the adjacency matrix $\mathcal{A}$. If cycles do not exist, the non-diagonal entries are set to $0$ and hence $G_{R_i}$ is an identity matrix. Otherwise, further analysis on the connections among data groups at each road segment is operated sequentially in the following three steps:

\subsubsection*{i.\,Node Analysis on Dual Graph}
The analysis starts with identifying road segments that have only outgoing flow, i.e., source nodes in the graph. The source nodes can be identified from the adjacency matrix $\mathcal{A}$ by checking the sum of each column. In particular, road segment $R_i$ is a source node when the sum of the $i^{\text{th}}$ column of $\mathcal{A}$ satisfies $\sum\limits_{l=1}^L a_{li} = 0,$ where $a_{li}$ is the ${(l,i)}^{\text{th}}$ entry of the adjacency matrix, which represents the edge $(R_l, R_i)$. 

\subsubsection*{ii.\,Trajectory Flow for Data Groups from Source Nodes}
If the road segment $R_i$ is a source node, the $m_i$ data groups in $R_i$ resulting from the MLKM method are considered to be $m_i$ unique targets. Then the trajectories of these $m_i$ targets are traced in the road network. In particular, if $\bar{T}_{iz} \cap  X_{iN_i} = \emptyset$, the target associated with data group $\bar{T}_{iz}$ does not contain any measurement from sensor $S_{iN_i}$, which corresponds to the case when target stops in the road segment or the data collection terminates before the target could approach to sensor $S_{iN_i}$. The trajectory tracking for this target is then completed. Otherwise, the permutation matrix $G_{i}$ of intersection ${i}$ that is consisted of sensor $S_{iN_i}$ is utilized to pinpoint the trajectory of the same target in the intersection, and its data group $\bar{T}_{lz}$ in the subsequent node or sink node $R_l$ where it is heading to. The trajectory tracking of the same target on the new road segments will keep on until the target stops or leaves the road network. The same process is used for tracing the flow of other targets.

\subsubsection*{iii.\,Matrix Description of Intermediate Nodes}
After the trajectories of all targets from the road segments have been confirmed, data points for each target on different road segments can be merged. More precisely, the corresponding entry $(p,q)$ in $G_{R_l}$ that is assigned as $1$ means that data groups $\bar{T}_{lp}$ and $\bar{T}_{lq}$ belong to one target. Consequently, the corresponding matrix $G_{R_l}$ can be determined.


\section{SIMULATION}\label{sec:simu}
In this section, the performance of the proposed G-MLKM algorithm is evaluated. We first introduce the testing datasets generation process. Then the performance of the MLKM method on one road segment is evaluated and compared with k-means++ and DNN. Then the complete G-MLKM algorithm performance is evaluated. A detailed example presenting the output via using G-MLKM is given to show how matrices $G_{i_{nts}}$ and $G_{R_i}$ are created for data pairing at intersections and group merging. 

\subsection{Testing Data Generation}
In order to obtain a quantitative performance evaluation of the data association techniques, labeled data is needed to obtain the percentage of true association between targets and their measurements. One convenient way to have accurate labeled dataset for data-target association is to generate it artificially. Let the generated testing dataset from the road network be $M_t = \{ T_1, T_2, \cdots, T_L\}$, where $T_i \in \mathbb{R}^{m_i \times m_i}$ has the same data structure as $\mathcal{T}_i$ defined in (\ref{eq: result for 1D}). In particular, each element in $T_i$ is a data group that belongs to one target. Moreover, for any $T_i$ collected from road segments that have both incoming and outgoing flows, multiple rows may belong to the same target. 

We utilize the road network structure shown in Fig. \ref{fig:road_network} as a prototype for testing data generation. Moreover, $N_S$ sensors are assumed to be equally distributed on each road segment, where the length of road segment is $N_S\times d$. The position set for sensors is selected as ${\mathcal{P}}_i =\{d, 2d, \cdots, N_Sd\}$ with respect to the starting point of road segment $R_i$. The intersections are considered to have the same radius with the value of $d/2$. Hence, the distance between any two adjacency sensors is $d$. To further simplify the data generation process, we assume road segment $R_1$ is the only entrance of the road network during the data collection period with the incoming targets number be $N_A$, and targets have equal possibilities of valid heading directions at each intersection. The targets are assumed to move with a constant velocity and the velocity is also discretely affected by Gaussian noise, such that, $v_{ij} = {v}_0 + \mathcal{N}(\mu, \sigma),$ where ${v}_{ij}$ is one velocity measurement at sensor $S_{ij}$, $v_0$ is the velocity measurement at the previous sensor. The corresponding time measurement is calculated as $ {t}_{ij} = {t}_0 + v_{ij}/{(j\cdot d)}.$ The initial velocity and time for the $N_A$ targets are uniformly selected from the range $(v_{min}, v_{max})$ and $(t_{min}, t_{max})$, respectively (refer Table \ref{tb:sim_parameters1}). The testing dataset generating process stops when all targets move out of the road network.

With the generated testing datasets, we may evaluate the performance of the data-target association techniques by calculating the data association accuracy, which is defined as the ratio between correctly classified number of data ($M_{cr}$) and the total number of data ($M_{t}$), such that, $\frac{\textup{numel}(M_{cr})}{\textup{numel}(M_{t})}\times100\%,$ where $\textup{numel}(M)$ returns the number of elements in $M$. As multiple testing datasets are generated, the provided statistical information about performance includes the minimum (left - blue bar), average (middle - orange bar), and maximum (right - yellow bar) accuracies.

\subsection{MLKM Performance and Comparisons}
Before evaluating the entire accuracy of the proposed G-MLKM algorithm, the MLKM method is evaluated and compared with the other two common data clustering machine learning techniques, in particular, k-means++ and DNN, based on the collected dataset in road segment $R_1$.

\subsubsection{K-means++} 
\begin{figure}[h]
	\centering
	\includegraphics[width=5cm]{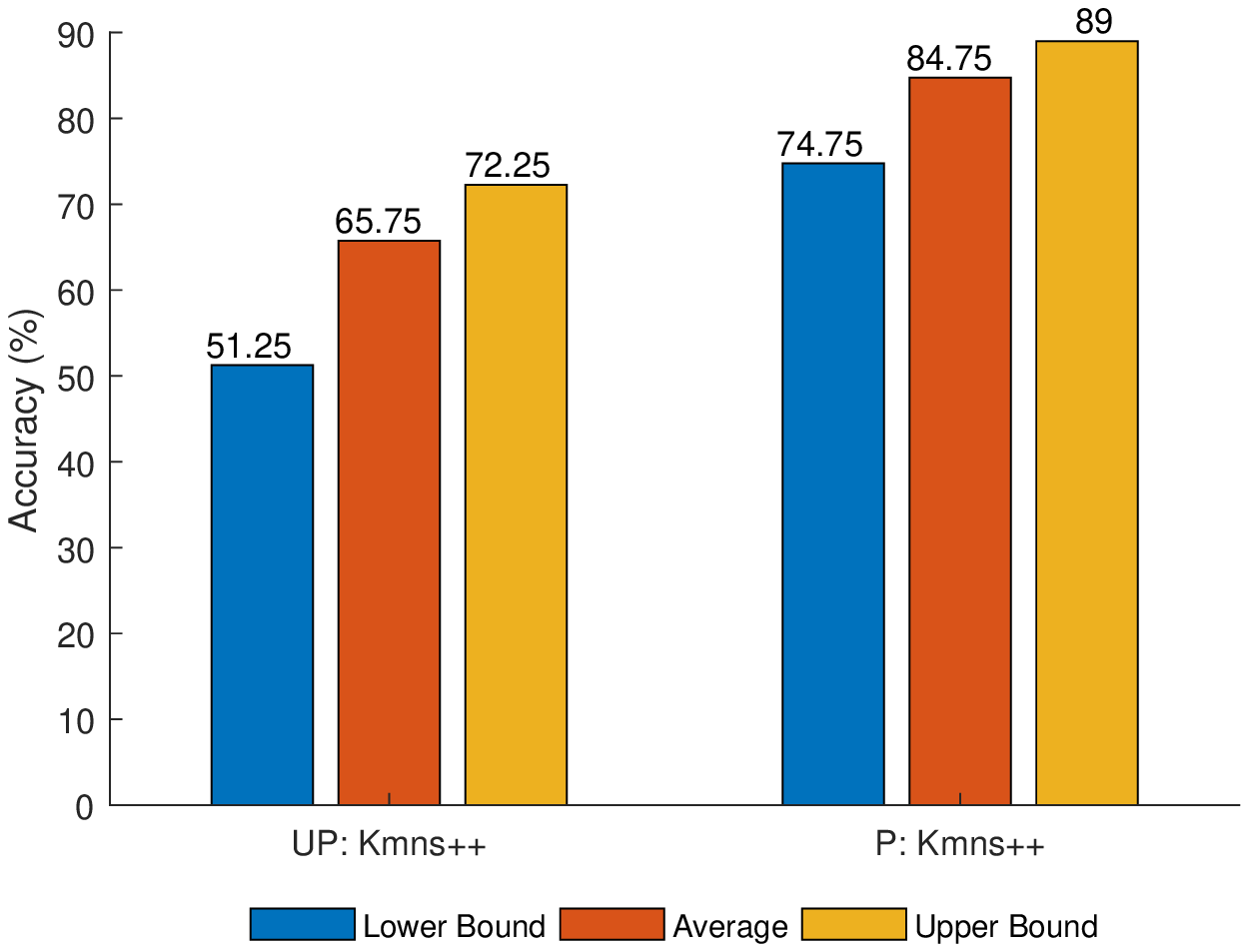}
	\caption{K-means++ accuracy for Simulation 1 parameters on unprocessed (UP) and preprocessed (P) data.}
	\label{fig:kmeanscomp10t10s}
\end{figure}

\begin{figure}[h]
	\centering
	\includegraphics[width=5cm]{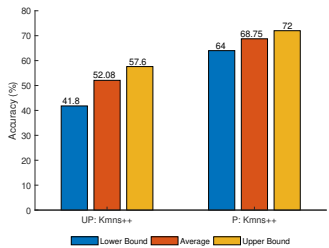}
	\caption{K-means++ accuracy for Simulation 2 parameters on unprocessed (UP) and preprocessed (P) data.}
	\label{fig:kmeanscomp50t20s}
\end{figure}

The first set of simulations evaluate the performance of K-means++ based on two criteria: (i) unprocessed vs. preprocessed data, and (ii) using different values of $N_A$ and $N_S$. When the values of $N_A$ and $N_S$ increase, more data points are introduced into the dataset, leading to more overlapping among these data points. Figures \ref{fig:kmeanscomp10t10s} and \ref{fig:kmeanscomp50t20s} show the performance of K-means++ using the parameters listed in Table \ref{tb:sim_parameters1}.

\begin{table} 
	\centering
	\caption{Simulation parameters.}
	\begin{tabular}{|c|c|c|}		
		\hline 
		& Simulation 1 & Simulation 2 \\ 
		\hline 
		$N_A$ & 10 & 50 \\ 
		\hline 
		$N_S$ & 10 & 20 \\ 
		\hline 
		$(v_{min},v_{max})$ & $U(10,50)$ & $U(10,50)$   \\ 
		\hline 
		$(t_{min},t_{max})$ & $U(0,40)$ & $U(0,40)$  \\ 
		\hline 
	\end{tabular}
    \label{tb:sim_parameters1}
\end{table}

As can be observed, a higher accuracy is achieved using the preprocessed data than that using the unprocessed data. This can be seen by comparing average, maximum and minimum accuracy for the two methods that use the preprocessed data versus the the unprocessed data, as shown in Figure \ref{fig:kmeanscomp10t10s}. Using the raw data, the measurements associated with a specific target are sparse along the time axis. However, the velocity measurements from the same sensor are closely grouped along the velocity axis. These conditions contribute to incorrect clustering of the data. The preprocessing technique reduces the distance between target related measurements, therefore reducing the effect of the velocity measurements on the clustering. 

A low accuracy is obtained for large values of $N_A$ and $N_S$. This can be observed by comparing average, maximum and minimum accuracy for different $N_A$ and $N_S$, as shown in Figures \ref{fig:kmeanscomp10t10s} and \ref{fig:kmeanscomp50t20s}. Similar to the unprocessed data, a large number of sensors/targets increases the density of measurement points. The concentration of measurements increases the probability that K-means/K-means++ clusters the data incorrectly (even with preprocessing).

\subsubsection{DNN} The K-means++ fails to correctly cluster data when overlapping of measurements occurs. Deep neural networks (DNN) is used as an alternative approach because it has been shown to provide good results to uncover patterns for large dataset classification. One necessary condition for DNN is the availability of labeled datasets for training. To meet the requirements of DNN, it is assumed that labeled data is available for training.

The results for DNN are obtained using $N_A= 50$ targets and $N_S=50$ sensors. Assuming that a portion of the data association has already been identified, the objective is to train a neural network to label the unidentified measurements. The number of `training' sensors that provide labeled information and `testing' sensors that provide unlabeled information are provided in Table \ref{tb:accuracy_nn}. The accuracy is obtained for various proportions of `training' sensors to `testing' sensors. Table \ref{tb:accuracy_nn} also shows the accuracy obtained for different dataset configuration.

\begin{table}[h]
	\centering
	\caption{DNN with different training and testing datasets.}
	\begin{tabular}{|c|c|c|c|}
		\hline 
		Train Sensors & Test Sensors & Train Accuracy & Test Accuracy \\ 
		\hline 
		20 & 30 & 98\% & 68\% \\ 
		\hline 
		25 & 25 & 97.8\% & 68\% \\ 
		\hline 
		30 & 20 & 99\% & 72\% \\ 
		\hline 
		40 & 10 & 98.6\% & 84.4\% \\ 
		\hline 
		45 & 5 & 98.9\% & 91.6\% \\ 
		\hline 
	\end{tabular} 
	\label{tb:accuracy_nn}%
\end{table}

It can be observed that the training (respectively, testing) accuracy is high (respectively, low), when the testing dataset is relatively small. However, when the testing dataset is relatively high, the testing performance increases significantly (up to 91\%). A high training accuracy with a low testing accuracy means that DNN suffers from overfitting due to the small size of training dataset. Given this comparison, DNN is applicable when a large portion of training dataset is available to train the network for classifying a relatively small amount of measurements.

\subsubsection{MLKM} K-means++ does not provide good accuracy for a high number of measurements but performs well when clustering small amounts of data. DNN can cluster large datasets but requires a large training dataset. MLKM combines the multi-layer back-propagation error correction from DNN and the clustering capabilities of K-means++. The DNN-inspired error correction significantly improves the performance of MLKM by preventing the clustering errors in layer 1 to propagate to the cluster association in layer 3. 

\begin{figure}[h]
	\centering
	\includegraphics[width=5cm]{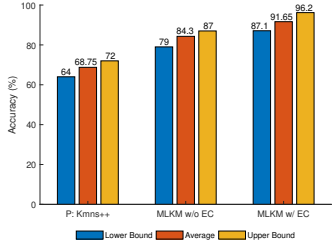}
	\caption{K-means++ for preprocessed data (P:K-means++), MLKM without error correction (MLKM w/o EC)  and MLKM with error correction (MLKM w/ EC).}
	\label{fig:ModelComp}
\end{figure}

The results for the MLKM method are obtained using $N_A= 50$ number of targets and $N_S=20$ number of sensors. In addition, the time and velocity parameters are set to $(t_{min},t_{max}) = \mathcal{U}(-10,30)$ and $(v_{min},v_{max}) = \mathcal{N}(50,40)$, receptively. Figure \ref{fig:ModelComp} shows the performance of the MLKM method with and without error correction, as well as results using the standard K-means++ method with preprocessing.

It can be observed that a higher accuracy is achieved using MLKM than that using K-means++. Figure \ref{fig:ModelComp} shows the average, maximum and minimum accuracy for both methods. The error correction performed in layer 2 improves the average accuracy of MLKM by approximately 7\% (MLKM w/ EC 91.65\%; MLKM w/o EC 84.3\%). 

\subsection{G-MLKM Overall Performance}
\begin{figure}[h!]
	\centering
	\includegraphics[width=5cm]{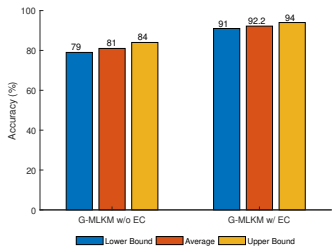}
	\caption{Accuracy Obtained for Extended MLKM w/ EC and w/o EC.}
	\label{fig:ModelComp2}
\end{figure}

The results for the G-MLKM method are obtained using $N_A= 20$ number of targets and $N_S=10$ number of sensors. In addition, the time and velocity parameters are set to $(t_{min},t_{max}) = U(0,40)$ and $(v_{min},v_{max}) = U(10,50)$, respectively. Figure \ref{fig:ModelComp2} shows the performance of the G-MLKM algorithm with and without error correction.

It can be observed that a higher accuracy is achieved using G-MLKM with error correction than the result without error correction. Figure \ref{fig:ModelComp2} shows the average, maximum and minimum accuracy for both methods. The second error correction performed in the algorithm improves the average accuracy of G-MLKM by approximately 11\% (G-MLKM w/ EC 92.2\%; G-MLKM w/o EC 81\%).

\subsection{Matrix Output of the G-MLKM Algorithm}
The proposed G-MLKM algorithm implements multiple (determined by the structure of road networks) permutation matrices $G_{i_{nts}}$ and $L$ symmetric matrices $G_{R_i}$ to represent the data cluster classification results at intersections and road segments, respectively. A detail example is illustrated to show the use of proposed G-MLKM matrix output.

Suppose 5 targets (named as $N_1, N_2, N_3, N_4, N_5$, respectively) go through the road network as shown in Fig. \ref{fig:road_network} during a certain time. The trajectory ground truth is listed in Table \ref{tb:ground-truth}. In particular, road segment $R_1$ has three data groups denoted as $\{1, 2, 3\}$, $R_2$ has six data groups denoted as $\{1, 2, 3, 4, 5, 6\}$, $R_3$ has five data groups denoted as $\{1, 2, 3, 4, 5\}$, $R_4$ has three data groups denoted as $\{1, 2, 3\}$, $R_5$ has one data groups denoted as $\{1\}$, and $R_6$ has two data groups denoted as $\{1, 2\}$. 

Take target $N_1$ as an example, it travels through road segment $R_1, R_2$, then heads to road segment $R_5$. After that, it keeps on moving through road segment $R_4, R_2$ and finally leaves the road network through road segment $R_3$. The connections among associated data groups in each road segment that are related to target $N_1$ is represented as $\{1^1 , {1^2} , {6^2} , 1^5 , 3^4 ,  5^3\},$ which means data group 1 in road segment $R_1$, data groups 1 and 6 in road segment $R_2$, data group 1 in road segment $R_5$, data group 3 in road segment $R_4$, and data group 5 in road segment $R_3$ all belong to the measurements extracted from target $N_1$.  

As the road segment $R_2$ has two data groups belong to one target, the ideal matrix $G_{R_2}\in \mathbb{R}^{6\times6}$ should be 
\begin{align*}
G_{R_2}=
\begin{bmatrix}
1 & 0 & 0 & 0 & 0 & 1 \\
0 & 1 & 0 & 0 & 0 & 0 \\
0 & 0 & 1 & 0 & 0 & 0 \\
0 & 0 & 0 & 1 & 0 & 0 \\
0 & 0 & 0 & 0 & 1 & 0 \\
1 & 0 & 0 & 0 & 0 & 1
\end{bmatrix},
\end{align*}
with respect to its data groups $\{1, 2, 3, 4, 5, 6\}$. For the other road segments, the corresponding matrix $G_{R_i}$ is an identity matrix related to its own data groups. Especially, $G_{R_1} = I^{3\times 3}$, $G_{R_3}= I^{3\times 3}$, $G_{R_4}= I^{3\times 3}$, $G_{R_5}= I^{1\times 1}$, $G_{R_6}= I^{2\times 2}$.

Let the intersection formed by road segments $R_6, R_5$ and $R_4$ be denoted as $a$. The incoming dataset $Q^{a}_I \in \mathbb{R}^{3\times 1}$ can be stored in the sequence of $\{1^5, 1^6, 2^6\}$ and the outgoing dataset $Q^{a}_O \in \mathbb{R}^{3\times 1}$ can be stored in the sequence of $\{1^4, 2^4, 3^4\}$. Therefore, the permutation matrix $G_a$ may determined as
\begin{align*}
G_{a}=
\begin{bmatrix}
0 & 0 & 1 \\
1 & 0 & 0 \\
0 & 1 & 0  
\end{bmatrix}.
\end{align*}
Similarly, for the intersection formed by road segment $R_1, R_2$ and $R_4$ (named as intersection $b$), matrix $G_b$ may determined as
\begin{align*}
G_{b}=
\begin{bmatrix}
1 & 0 & 0 & 0 & 0 & 0 \\
0 & 1 & 0 & 0 & 0 & 0 \\
0 & 0 & 1 & 0 & 0 & 0 \\
0 & 0 & 0 & 1 & 0 & 0 \\
0 & 0 & 0 & 0 & 1 & 0 \\
0 & 0 & 0 & 0 & 0 & 1
\end{bmatrix},
\end{align*}
with $Q^{b}_I \in \mathbb{R}^{6\times 1}$ stored in the sequence of $\{1^1$, $2^1$, $3^1$, $1^4$, $2^4$, $3^4\}$ and the outgoing dataset $Q^{b}_O \in \mathbb{R}^{6\times 1}$ in the sequence of $\{1^2$, $2^2$, $3^2$, $4^2$, $5^2$, $6^2\}$. For the intersection formed by road segment $R_2, R_3$ and $R_5$ (named as intersection $c$), $G_c$ may determined as
\begin{align*}
G_{c}=
\begin{bmatrix}
0 & 0 & 0 & 0 & 0 & 1 \\
1 & 0 & 0 & 0 & 0 & 0 \\
0 & 1 & 0 & 0 & 0 & 0 \\
0 & 0 & 1 & 0 & 0 & 0 \\
0 & 0 & 0 & 1 & 0 & 0 \\
0 & 0 & 0 & 0 & 1 & 0
\end{bmatrix},
\end{align*}
with $Q^{c}_I \in \mathbb{R}^{6\times 1}$ stored in the sequence of $\{1^2$, $2^2$, $3^2$, $4^2$, $5^2$, $6^2\}$ and the outgoing dataset $Q^{c}_O \in \mathbb{R}^{6\times 1}$ in the sequence of $\{1^3$, $2^3$, $3^3$, $4^3$, $5^3$,$1^5\}$.

With these matrices determined, the output result from G-MLKM can be clearly presented.

\begin{table} 
	\centering
	\begin{tabular}{|c|c|c|}		
		\hline 
		Target & Trajectory & Representation \\ 
		\hline 
		$N_1$ & $\makecell{R_1 \rightarrow R_2 \rightarrow R_5\\ \rightarrow R_4 \rightarrow  R_2 \rightarrow  R_3}$ & $\{1^1, \textcolor{black}{1^2}, 1^5, 3^4 , \textcolor{black}{6^2} ,  5^3\}$  \\ 
		\hline 
		$N_2$ & $R_1 \rightarrow R_2 \rightarrow R_3$ & $\{2^1 , 2^2 , 1^3\}$  \\ 
		\hline 
		$N_3$ & $R_1 \rightarrow R_2 \rightarrow R_3$ & $\{3^1 , 3^2 , 2^3\}$   \\ 
		\hline 
		$N_4$ & $R_6 \rightarrow R_4 \rightarrow R_2 \rightarrow R_3$ & $\{1^6 ,1^4 , 4^2 , 3^3\}$   \\ 
		\hline 
        $N_5$ & $R_6 \rightarrow R_4 \rightarrow R_2 \rightarrow R_3$ & $\{2^6 , 2^4 , 5^2 , 4^3\}$   \\ 
        \hline
	\end{tabular}
    \caption{Ground Truth for 5 Targets Trajectories. The representation of $A^i$ denotes the associated data group $A$ in road segment $R_i$.}
    \label{tb:ground-truth}
\end{table}

\section{Conclusions and Future Work}\label{sec:con}
This paper has studied data pattern recognition for multi-targets in a constrained space, where the data is the minimal information provided by spatially distributed sensors. In contrast to the existing methods that rely on probabilistic hypothesis estimation, we proposed to utilize the machine learning approach for the data correlation analysis. Two common data clustering algorithms, namely, K-means++ and deep neural network, were first analyzed for data association given a simplified constrained space. Then the MLKM method was proposed via leveraging the structure advantage of DNN and the unsupervised clustering capability of k-means++. After that, graph theory was introduced in the purpose of extending the scope of MLKM for a general constrained space. In particular, we proposed a p-dual graph for data association at intersections and merged the results from local spaces and intersections through the dual graph of the constrained space. Simulation studies were provided to demonstrate the performance of the MLKM method and the proposed G-MLKM. Our future work will focus on releasing the assumptions in this paper to improve G-MLKM in the scenarios of false alarms.







\bibliography{refs} 
\bibliographystyle{ieeetr}

\end{document}